\documentclass[a4paper]{llncs}

\usepackage{wrapfig}
\usepackage{booktabs}
\usepackage{paralist}
\usepackage{xspace}
\usepackage{amsmath,amssymb,amsfonts}
\usepackage{url}
\usepackage{algorithm}
\usepackage[noend]{algpseudocode}
\usepackage{subcaption}
\usepackage{placeins}

\usepackage{hyperref}
\hypersetup{colorlinks,breaklinks,urlcolor=blue,linkcolor=blue,citecolor=blue}

\newcommand{\relu}{\text{ReLU}\xspace{}}
\newcommand{\sat}{\texttt{SAT}}
\newcommand{\unsat}{\texttt{UNSAT}}
\makeatletter
\newcommand{\setlabel}[2]{%
  \phantomsection
  #1\def\@currentlabel{\unexpanded{#1}}\label{#2}%
}
\makeatother

\newcommand{\tmax}{T_{\max}}
\newcommand{\snapshot}{\mathcal{S}}

\newcommand{\mysubsubsection}[1]{\medskip\noindent\textbf{#1}}
\usepackage[dvipsnames]{xcolor}

\usepackage{eqparbox}

\usepackage{tikz} \usetikzlibrary{arrows,automata, matrix}
\usetikzlibrary{arrows,trees,backgrounds,automata,shapes,decorations,plotmarks,fit,calc,positioning,shadows,chains,shapes.geometric}
\tikzstyle{every pin edge}=[<-,shorten <=1pt]
\tikzstyle{neuron}=[circle,fill=black!25,minimum size=17pt,inner sep=0pt]
\tikzstyle{input neuron}=[neuron, fill=orange!50]
\tikzstyle{output neuron}=[neuron, fill=purple!50]
\tikzstyle{hidden neuron}=[neuron, fill=blue!50]
\tikzstyle{shadow neuron}=[neuron, regular polygon, regular polygon sides=4,fill=blue!30]
\tikzstyle{new neuron}=[neuron, fill=gray!50]
\tikzstyle{annot} = [text width=3.5em, text centered]
\tikzstyle{nnedge} = [-{stealth},shorten >=0.1cm, shorten <=0.05cm,line width=0.8pt,black]

\begin{document}

\title{Verifying Recurrent Neural Networks using Invariant Inference}

\author{}
\institute{}
\author{
  Yuval Jacoby\inst{1} \and
  Clark Barrett\inst{2} \and
  Guy Katz\inst{1} 
}
\institute{
  The Hebrew University of Jerusalem, Israel \\
  \{yuval.jacoby, g.katz\}@mail.huji.ac.il
  \and
  Stanford University, USA \\
  clarkbarrett@stanford.edu 
}

\maketitle

\begin{abstract}
  Deep neural networks are revolutionizing the way complex systems are
  developed. However, these automatically-generated networks are 
  opaque to humans, making it difficult to reason about them and
  guarantee their correctness. Here, we propose a novel approach for
  verifying properties of a widespread variant of neural networks,
  called \emph{recurrent neural networks}. Recurrent neural networks
  play a key role in, e.g., speech recognition, and their
  verification is crucial for guaranteeing the reliability of many
  critical systems. Our approach is based on the inference of
  \emph{invariants}, which allow us to reduce the complex problem of
  verifying recurrent networks into simpler, non-recurrent
  problems. Experiments with a proof-of-concept implementation of our
  approach demonstrate that it performs 
  orders-of-magnitude better than the state of the art.
\end{abstract}

\section{Introduction}
The use of \emph{recurrent neural networks} (\emph{RNN}s)~\cite{dnn}
is on the rise. RNNs are a particular kind of deep neural networks
(DNNs), with the useful ability to store information from previous
evaluations in constructs called \emph{memory units}. This
differentiates them from \emph{feed-forward neural networks}
(\emph{FFNNs}), where each evaluation of the network is performed
independently of past evaluations.  The presence of memory units
renders RNNs particularly suited for tasks that involve context, such
as machine translation~\cite{DeChLeTo18}, health
applications~\cite{LiKaElWe16}, speaker recognition~\cite{WaWaPaLo18},
and many other tasks where the network's output might be affected by
previously processed inputs.

Part of the success of RNNs (and of DNNs in general) is attributed to
their very attractive generalization properties: after being trained
on a finite set of examples, they generalize well to inputs they have
not encountered before~\cite{dnn}. Unfortunately, it is known that
RNNs may react in highly undesirable ways to certain inputs. For
instance, it has been observed that many RNNs are vulnerable to
\emph{adversarial inputs}~\cite{CiAdNeKe17,SzZaSuBrErGoFe13}, where
small, carefully-crafted perturbations are added to an input in order
to fool the network into a classification error. This example, and
others, highlight the need to \emph{formally verify} the correctness
of RNNs, so that they can reliably be deployed in safety-critical
settings. However, while DNN verification has received significant
attention in recent years
(e.g.,~\cite{BuTuToKoMu18,ChNuHuRu18,ChNuRu17,Eh17,GeMiDrTsChVe18,GoAdKeKa20,HuKwWaWu17,KaBaDiJuKo17,Marabou,LoMa17,NaKaRySaWa17,TjXiTe19,WaPeWhYaJa18,WuOzZeIrJuGoFoKaPaBa20}),
almost all of these efforts have been focused on FFNNs, with very
little work done on RNN verification.

To the best of our knowledge, the only existing general approach for
RNN verification is via \emph{unrolling}~\cite{AkKeLoPi19}: the RNN is
duplicated and concatenated onto itself, creating an equivalent
feed-forward network that operates on a sequence of $k$ inputs
simultaneously, as opposed to one at a time. The FFNN can then be
verified using existing verification technology. The
main limitation of this approach is that unrolling
increases the network size by a factor of $k$ (which, in real-world applications,
can be in the hundreds~\cite{WaWaPaLo18}).
Because the complexity of FFNN verification is known to
be worst-case exponential in the size of the network~\cite{ReluPlex}, this reduction gives
rise to FFNNs that are difficult to verify --- and is hence applicable
primarily to small RNNs with short input sequences.

Here, we propose a novel approach for RNN verification, which affords
far greater scalability than unrolling.  Our approach also
reduces the RNN verification problem into FFNN verification, but does
so in a way that is independent of the number of inputs that the RNN
is to be evaluated on. Specifically, our approach consists of two main
steps: (i) create an FFNN that \emph{over-approximates} the RNN, but
which is the same size as the RNN; and (ii) verify properties over this
over-approximation using existing techniques for FFNN verification.
Thus, our approach circumvents any duplication of the network or its
inputs.

In order to perform step (i), we leverage the well-studied notion of
\emph{inductive invariants}: our FFNN encodes time-invariant
properties of the RNN, which hold initially and continue to hold after
the RNN is evaluated on each additional input. Automatic inference of
meaningful inductive invariants has been studied extensively
(e.g.,~\cite{NgAnRuHi17,ShDiDiAi11,SiDaRaNaSo18}), and is known to be
highly difficult~\cite{PaImShKaSa16}. We propose here an approach for
generating invariants according to \emph{predefined templates}. By
instantiating these templates, we automatically generate a candidate
invariant $I$, and then: (i) use our underlying FFNN verification
engine to prove that $I$ is indeed an invariant; and (ii) use $I$ in
creating the FFNN over-approximation of the RNN, in order to prove the
desired property. If either of these steps fail, we refine $I$ (either
strengthening or weakening it, depending on the point of failure), and
repeat the process. The process terminates when the property is proven
correct, when a counter-example is found, or when a certain timeout
value is exceeded.

We evaluate our approach using a proof-of-concept implementation,
which uses the Marabou tool~\cite{Marabou} as its FFNN verification
back-end. When compared to the state of the art on a set of benchmarks
from the domain of speaker recognition~\cite{WaWaPaLo18}, our approach
is orders-of-magnitude faster. Our implementation, together with our benchmarks
and experiments, is available online~\cite{RNNCode}.
     
The rest of this paper is organized as follows.  In
Sec.~\ref{sec:background}, we provide a brief background on DNNs and
their verification. In Sec.~\ref{sec:method}, we describe our approach
for verifying RNNs via reduction to FFNN verification, using
invariants. We describe automated methods for RNN invariant inference
in Sec.~\ref{sec:invariant_generation}, followed by an evaluation of
our approach in Sec.~\ref{sec:evaluation}. We then discuss related
work in Sec.~\ref{sec:relatedWork}, and conclude with
Sec.~\ref{sec:conclusion}.

\section{Background}
\label{sec:background}

\subsection{Feed-Forward Neural Networks and their Verification}

An FFNN $N$ with $n$ layers consists of an input layer, $n-2$ hidden
layers, and an output layer.
We use $s_i$ to denote the \emph{dimension} of layer $i$, which is the number of 
neurons in that layer. We use $v_{i,j}$ to refer to the $j$-th neuron in the
$i$-th layer.
Each hidden layer is associated with a weight matrix $W_i$ and a bias
vector $b_i$. The FFNN input vector is denoted as $v_1$, and the output
vector of each hidden layer $1 < i < n$ is $v_i = f\left( W_i v_{i-1} + b_i
\right)$, where $f$ is some element-wise activation function (such as
$\relu{}(x) = \max{}(0,x)$). The output layer is evaluated similarly, but
without an activation function: $v_n = W_{n-1}v_{n-1}+b_n$.  Given an input
vector $v_1$, the network is evaluated by sequentially calculating $v_i$ for
$i=2,3,\ldots,n$, and returning $v_n$ as the network's output.

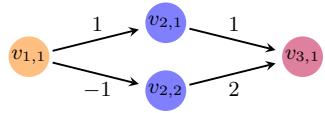
\begin{wrapfigure}[8]{r}{4.5cm}
  \vspace{-1cm}
  \begin{center}
    \scalebox{0.9}{
      \def\layersep{2.0cm}
\begin{tikzpicture}[shorten >=1pt,->,draw=black!50, node distance=\layersep,font=\footnotesize]

  \node[input neuron] (I-1) at (0,-1) {$v_{1,1}$};

  \path[yshift=0.5cm] node[hidden neuron] (H-1)
  at (\layersep,-1 cm) {$v_{2,1}$};
  \path[yshift=0.5cm] node[hidden neuron] (H-2)
  at (\layersep,-2 cm) {$v_{2,2}$};

  \node[output neuron] at (2*\layersep, -1) (O-1) {$v_{3,1}$};

  \draw[nnedge] (I-1) -- node[above] {$1$} (H-1); 
  \draw[nnedge] (I-1) -- node[below] {$-1$} (H-2);
  \draw[nnedge] (H-1) -- node[above] {$1$} (O-1);
  \draw[nnedge] (H-2) -- node[below] {$2$} (O-1); 
  
\end{tikzpicture}

    }
    \caption{A simple feed-forward neural network.}
    \label{fig:fullyConnectNetwork}
  \end{center}
\end{wrapfigure}

A simple example appears in Fig.~\ref{fig:fullyConnectNetwork}. This
FFNN has a single input neuron $v_{1,1}$, a single
output neuron $v_{3,1}$, and two hidden neurons $v_{2,1}$ and
$v_{2,2}$. All bias values are assumed to be
$0$, and we use the common $\relu{}(x) = \max(0,x)$ function as our activation function.
When the input neuron is assigned
$v_{1,1}=4$, the weighted sum and activation functions
yield $v_{2,1}=\relu{}(4)=4$ and $v_{2,2}=\relu{}(-4)=0$. Finally, we obtain
the output $v_{3,1} = 4$.

\mysubsubsection{FFNN Verification.}
In FFNN verification we seek inputs that satisfy certain constraints,
such that their corresponding outputs also satisfy certain
constraints. Looking again at the network from
Fig.~\ref{fig:fullyConnectNetwork}, we might wish to know whether
$v_{1,1}\leq 5$ always entails $v_{3,1}<20$. Negating the output
property, we can use a verification engine to check whether it is
possible that $v_{1,1}\leq 5$ and $v_{3,1}\geq 20$. If this query is
unsatisfiable (\unsat{}), then the original property holds; otherwise,
if the query is satisfiable (\sat{}), then the verification engine
will provide us with a counter-example (e.g.,
$v_{1,1}=-10, v_{3,1}=20$ in our case).

Formally, we define an FFNN verification query as a triple
$\langle P,N,Q\rangle$, where $N$ is an FFNN, $P$ is a predicate over
the input variables $x$, and $Q$ is a predicate over the output
variables $y$. Solving this query entails deciding whether there
exists a specific input assignment $x_0$ such that
$P(x_0)\wedge Q(N(x_0))$ holds (where $N(x_0)$ is the output of $N$ for the
input $x_0$). It has been shown that even for simple FFNNs and for predicates
$P$ and $Q$ that are conjunctions of linear constraints, the
verification problem is NP-complete~\cite{ReluPlex}: in the
worst-case, solving it requires a number of operations that is 
exponential in the number of neurons in $N$.

\subsection{Recurrent Neural Networks}

Recurrent Neural Networks (RNNs) are similar to FFNNs, but have an
additional construct called a \emph{memory unit}. Memory units allow a hidden
neuron to \emph{store} its assigned value for a specific evaluation
of the network, and have that value become part of the neuron's
weighted sum computation in the \emph{next} evaluation. Thus, when
evaluating the RNN in time step $t+1$, e.g. when the RNN reads the
$t+1$'th word in a sentence, the results of the $t$ previous evaluations can
affect the current result. 

\begin{figure}[htp]
  \begin{minipage}{.42\textwidth}
    \begin{center}
      \begin{tabular}{cc|ccccc}
        \toprule
        Time Step &&&
        $v_{1,1}$ &
        $v_{2,1}$ &
        $\tilde{v}_{2,1}$ &
        $v_{3,1}$ 
        \\
        \midrule
        1 &&& 0.5 & 0.5 & 0 & 0.5 \\
        2 &&& 1.5 & 2 & 0.5 & 2 \\
        3 &&& -1 & 1 & 2 & 1 \\
        4 &&& -3 & 0 & 1 & 0 \\
        \bottomrule
      \end{tabular}
    \end{center}
  \end{minipage}
  \begin{minipage}{.42\textwidth}
    \begin{center}
      \begin{tikzpicture}[shorten >=1pt,->,draw=black!50, node distance=\layersep,font=\footnotesize]
        \def\layersep{2.5cm}
        \node[input neuron] (I-1) at (0,-1) {$v_{1,1}$};
        \path[] node[shadow neuron] (H-1-s)  at (0.5*\layersep,0.5) {$\tilde{v}_{2,1}$};
        \path[] node[hidden neuron] (H-1)  at (\layersep,-1) {$v_{2,1}$};
        \node[output neuron] at (2*\layersep, -1) (O-1) {$v_{3,1}$};
        
        \draw[nnedge] (I-1) -- node[above] {$1$} (H-1);
        \draw[nnedge] (H-1-s) edge[bend left, black] node[above] {$\ 1$} (H-1);
        \draw[nnedge] (H-1) edge[bend left, dotted, black] node[] {} (H-1-s);
        \draw[nnedge] (H-1) -- node[above] {$1$} (O-1); 
        
      \end{tikzpicture}
    \end{center}
  \end{minipage}
  \caption{ An illustration of a toy RNN with the ReLU activation
    function. Each row of the table represents a single time step,
    and depicts the value of each neuron for that step. Using a $t$
    superscript to represent time step $t$, we observe that
    $v^t_{2,1}$ is computed as
    $\max{}(0, \tilde{v}^{t}_{2,1} + v^t_{1,1})$, according to the ReLU
    function. }
  \label{fig:rnn_1d}
\end{figure}
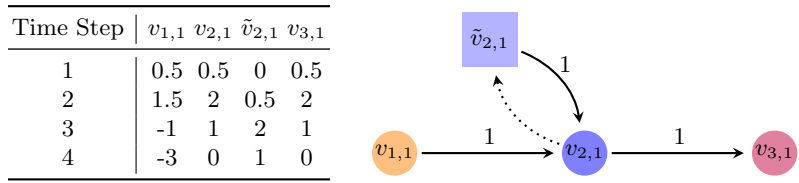

A simple RNN appears in Fig.~\ref{fig:rnn_1d}. There, node $\tilde{v}_{2,1}$
represents node $v_{2,1}$'s memory unit (we draw memory units as squares, and
mark them using the tilde sign). When computing the weighted sum for node
$v_{2,1}$, the value of $\tilde{v}_{2,1}$ is also added to the sum, according to
its listed weight ($1$, in this case). We then update $\tilde{v}_{2,1}$ for the
next round, using the vanilla RNN update rule: $\tilde{v}_{2,1} :=
v_{2,1}$. Memory units are initialized to $0$ for the first evaluation, at time
step $t=1$.

The FFNN definitions are extended to RNNs as follows. We use the $t$
superscript to indicate the timestamp of the RNN's evaluation: e.g.,
$v_{3,2}^4$ indicates the value that node $v_{3,2}$ is assigned in the
$4$'th evaluation of the RNN.  We associate each hidden layer of the
RNN with a square matrix $H_i$ of dimension $s_i$, which represents
the weights on edges from memory units to neurons.  Observe that each
memory unit in layer $i$ can contribute to the weighted sums of all
neurons in layer $i$, and not just to the neuron whose values it
stores.  For time step $t>0$, the evaluation of each hidden layer
$1<i<n$ is now computed by
$v^{t}_i = f\left(W_i v^{t}_{i-1} + H_i \tilde{v}^{t}_i + b_i\right)$,
and the output values are given by $v^t_n = W_nv^t_{n-1} + H_nv^{t-1}_n + b_n$.
By convention, we initialize memory units to $0$ (i.e. for every
memory unit $\tilde{v}$, $\tilde{v}^1 = 0$).  For simplicity, we
assume that each hidden neuron in the network has a memory unit. This
definition captures also ``regular'' neurons, by setting the
appropriate entries of $H$ to 0.

While we focus here on vanilla RNNs, our technique could be extended
to, e.g., LSTMs or GRUs; we leave
this for future work.

\mysubsubsection{RNN Verification.}
 We define an RNN verification query as a
tuple $\langle P, N, Q, \tmax\rangle$, where $P$ is an input property, $Q$ is
an output property, $N$ is an RNN, and $\tmax\in \mathbb{N}$ is a bound on
the time interval for which the property should hold. $P$ and $Q$ include linear
constraints over the network's inputs and outputs, and may 
also use the notion of time.

As a \setlabel{running example}{par:rnn_example},
consider the network from Fig.~\ref{fig:rnn_1d}, denoted by $N$, the
input predicate
$P = \bigwedge_{t=1}^5(-3\leq v^t_{1,1}\leq 3)$, the
output predicate
$Q = \bigvee_{t=1}^5(v_{3,1}^t \geq 16)$, and the time bound $\tmax=5$. This query searches for an
evaluation of $N$ with $5$ time steps, in which all input values are
in the range $[-3,3]$, and such that at some time step the output value is at least 16.
By the weights of $N$, it can be proved that $v^t_{3,1}$ is
at most the sum of the ReLUs of inputs so far, $v^t_{3,1}\leq
\sum_{i=1}^t\relu{}(v_{1,1}^i)\leq 3t$; and so $v_{3,1}^t\leq
15$ for all $1\leq t \leq 5$, and the query is \unsat{}.

\subsection{Inductive Invariants} 
\label{sec:inductive_invariant}
Inductive invariants~\cite{PaImShKaSa16} are a well-established way to reason
about software with loops. Formally, let $\langle Q, q_0, T\rangle$ be
a transition system, where $Q$ is the set of states, $q_0\in Q$ is an
initial state, and $T\subseteq Q\times Q$ is a transition relation.
An invariant $I$ is a logical formula defined over the states of $Q$,
with two properties:
\begin{inparaenum}[(i)]
\item $I$ holds for the initial state, i.e. $I(q_0)$ holds; and
\item $I$ is closed under $T$, i.e. $(I(q) \wedge
  \langle q,q' \rangle\in T) \Rightarrow I(q')$.
  \end{inparaenum}
  If it can be proved (in a given proof system) that formula $I$ is an
  invariant, we say that $I$ is an inductive invariant.

  Invariants are particularly useful when attempting to verify that a
  given transition system satisfies a \emph{safety property}. There,
  we are given a set of bad states $B$, and seek to prove that none of
  these states is reachable. We can do so by showing that
  $\{q\in Q \ |\ I(q)\}\cap B=\emptyset$.  Unfortunately,
  automatically discovering invariants for which the above holds is
  typically an undecidable problem~\cite{PaImShKaSa16}. Thus, a common
  approach is to restrict the search space --- i.e., to only search
  for invariants with a certain syntactic form.

\section{Reducing RNN Verification to FFNN Verification} 
\label{sec:method}

\subsection{Unrolling}

\begin{figure}[htp]
    \begin{center}
        \scalebox{0.8}{ \begin{tikzpicture}[shorten >=1pt,->,draw=black!50, node distance=\layersep,font=\footnotesize]
  \def\layerseph{1.5cm}
  \def\layersepv{-1.5cm}
    \node[input neuron] (I-1) at (0,0) {$v^{1}_{1,1}$};
    \node[input neuron] (I-2) at (1*\layerseph,0) {$v^{2}_{1,1}$};
    \node[input neuron] (I-3) at (2*\layerseph,0) {$v^{3}_{1,1}$};
    \node[input neuron] (I-4) at (3*\layerseph,0) {$v^{4}_{1,1}$};
    \node[input neuron] (I-5) at (4*\layerseph,0) {$v^{5}_{1,1}$};

    \node[hidden neuron]         (B) at (0,\layersepv) {$v^{1}_{2,1}$};
    \node[hidden neuron]         (C) at (1*\layerseph, \layersepv) {$v^{2}_{2,1}$};
    \node[hidden neuron]         (D) at (2*\layerseph, \layersepv) {$v^{3}_{2,1}$};
    \node[hidden neuron]         (E) at (3*\layerseph, \layersepv) {$v^{4}_{2,1}$};
    \node[hidden neuron]         (F) at (4*\layerseph, \layersepv) {$v^{5}_{2,1}$};

    \node[output neuron]         (O-1) at (0,2*\layersepv) {$v^{1}_{3,1}$};
    \node[output neuron]         (O-2) at (1*\layerseph, 2*\layersepv) {$v^{2}_{3,1}$};
    \node[output neuron]         (O-3) at (2*\layerseph, 2*\layersepv) {$v^{3}_{3,1}$};
    \node[output neuron]         (O-4) at (3*\layerseph, 2*\layersepv) {$v^{4}_{3,1}$};
    \node[output neuron]         (O-5) at (4*\layerseph, 2*\layersepv) {$v^{5}_{3,1}$};

    \draw[nnedge] (I-1) -- node[left] {$1$} (B);
    \draw[nnedge] (I-2) -- node[left] {$1$} (C);
    \draw[nnedge] (I-3) -- node[left] {$1$} (D);
    \draw[nnedge] (I-4) -- node[left] {$1$} (E);
    \draw[nnedge] (I-5) -- node[left] {$1$} (F);

    \draw[nnedge] (B) -- node[left] {$1$} (O-1);
    \draw[nnedge] (C) -- node[left] {$1$} (O-2);
    \draw[nnedge] (D) -- node[left] {$1$} (O-3);
    \draw[nnedge] (E) -- node[left] {$1$} (O-4);
    \draw[nnedge] (F) -- node[left] {$1$} (O-5);
    
    \draw[nnedge,red] (B) -- node[above] {$1$} (C);
    \draw[nnedge,red] (C) -- node[above] {$1$} (D);
    \draw[nnedge,red] (D) -- node[above] {$1$} (E);
    \draw[nnedge,red] (E) -- node[above] {$1$} (F);
\end{tikzpicture} }
        \caption{ Unrolling the network from Fig.~\ref{fig:rnn_1d}, for $\tmax=5$
            time steps. The edges in red fill the role of the
            memory units of the original RNN. The number of neurons in
            the unrolled network is $5$ times the number of neurons in
            the original. } 
    \label{fig:unroll}
    \end{center}
\end{figure}
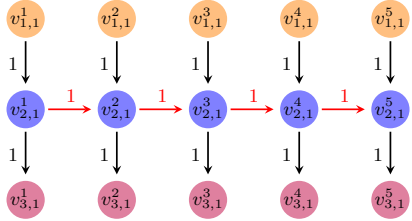

To date, the only available general approach for verifying
RNNs~\cite{AkKeLoPi19} is to transform the RNN in question into a
\emph{completely equivalent}, feed-forward network, using
\emph{unrolling}. An example appears in Fig.~\ref{fig:unroll}. The
idea is to leverage $\tmax$, which is an upper bound on the number of times
that the RNN will be evaluated. The RNN is duplicated $\tmax$ times, once
for each time step in question, and its memory units are
removed. Finally, the nodes in the $i$'th copy are used to fill the
role of memory units for the $i+1$'th copy of the network.

While unrolling gives a sound reduction from RNN verification to FFNN
verification, it unfortunately tends to produce very large networks.
When verifying a property that involves $t$ time steps, an RNN network
with $n$ memory units will be transformed into an FFNN with
$(t-1)\cdot n$ new nodes. Because the FFNN verification problem
becomes exponentially more difficult as the network size
increases~\cite{ReluPlex}, this renders the problem infeasible for
large values of $t$. As scalability is a major limitation of existing
FFNN verification technology, unrolling can currently only be applied
to small networks that are evaluated for a small number of time steps.

\subsection{Circumventing Unrolling}
\label{sec:overApproximation}

We propose a novel alternative to unrolling, which can reduce RNN
verification to FFNN verification without the blowup in network
size. The idea is to transform a verification query
$\varphi = \langle P, N, Q, \tmax\rangle$ over an RNN $N$ into a
different verification query
$\hat{\varphi} = \langle \hat{P}, \hat{N}, \hat{Q}\rangle$ over an
FFNN $\hat{N}$.  $\hat{\varphi}$ is not equivalent to $\varphi$, but
rather \emph{over-approximates} it: it is constructed in a way that
guarantees that if $\hat{\varphi}$ in \unsat{}, then $\varphi$ is also
\unsat{}. As is often the case, if $\hat{\varphi}$ is \sat{}, either
the original property truly does not hold for $N$, or the
over-approximation is too coarse and needs to be refined; we discuss
this case later.

A key point in our approach is that $\hat{\varphi}$ is created 
in a way that captures the notion of time in the FFNN setting, and
without increasing the network size. This is done by incorporating
into $\hat{P}$ an \emph{invariant} that puts bounds on the memory
units as a function of the time step $t$. This invariant does not
precisely compute the values of the memory units --- instead, it
bounds each of them in an interval. This inaccuracy is what makes
$\hat{\varphi}$ an over-approximation of $\varphi$. More specifically,
the construction is performed as follows:
\begin{enumerate}
  \item $\hat{N}$ is constructed from $N$ by adding a new input
    neuron, $t$, to represent time. In line with standard FFNN
    definitions, $t$ is treated as a
    real number.
  \item For every node $v$ with memory unit $\tilde{v}$, in $\hat{N}$
    we replace $\tilde{v}$ with a regular neuron, $v^m$, which is
    placed in the input layer.  Neuron $v^m$ will be connected to the
    network's original neurons with the original weights, just as
    $\tilde{v}$ was.\footnote{Note that we slightly abuse the
      definitions from Sec.~\ref{sec:background}, by allowing an input
      neuron to be connected to neurons in layers other than its
      preceding layer.}
  \item We set $\hat{P}=P\wedge (1\leq t\leq \tmax)\wedge I$, where
    $I$ is a formula that bounds the values of each new $v^m$ node as a function of
    the time step $t$. The constraints in $I$ constitute
    the invariant over the memory units' values.
    \item The output property is unchanged: $\hat{Q}=Q$.
\end{enumerate}

We name $\hat{\varphi}$ and $\hat{N}$ constructed in this way the
\emph{snapshot query} and the \emph{snapshot network}, respectively,
and denote  $\hat{\varphi} = \snapshot{}(\varphi)$ and $\hat{N}=\snapshot{}(N)$.
The intuition behind this construction is that query $\hat{\varphi}$
encodes a snapshot (an assignment of $t$) in which all constraints
are satisfied. At this point in time, the $v^m$ nodes
represent the values stored in the memory units (whose assignments
are bounded by the invariant $I$); and the input and output nodes
represent the network's inputs and outputs at time $t$. Clearly, a
satisfying assignment for $\hat{\varphi}$ does not necessarily
indicate a counter-example for $\varphi$; e.g., because the values
assigned to $v^m$ might be impossible to obtain at time $t$ in the
original network. However, if $\hat{\varphi}$ is \unsat{} then so is
$\varphi$, because there does not exist a point in time in which the
query might be satisfied. Note that the construction only increases the network size
by $1$ (the $v^m$ neurons replace the memory units, and we add a
single neuron $t$).

\mysubsubsection{Time-Agnostic Properties.}
In the aforementioned construction of $\hat{\varphi}$, the original
properties $P$ and $Q$ appear, either fully or as a conjunct, in the
new properties $\hat{P}$ and $\hat{Q}$. It is not immediately clear that this is
possible, as $P$ and $Q$ might also involve time. For example,
if $P$ is the formula $v_{1,2}^7\geq 10$, it cannot be
directly incorporated into $\hat{P}$, because $\hat{N}$ has no notion
of time step $7$.

For simplicity, we assume that $P$ and $Q$ are \emph{time-agnostic},
i.e. are given in the following form:
$P = \bigwedge_{t=1}^{\tmax}\psi_1$ and
$Q=\bigvee_{t=1}^{\tmax}\psi_2$, where $\psi_1$ and $\psi_2$ 
contain linear constraints over the inputs and outputs of $N$,
respectively, at time stamp $t$.  This formulation can express queries
in which the inputs are always in a certain interval, and a bound
violation of the output nodes is sought. Our running example from
Fig.~\ref{fig:rnn_1d} has this structure. When the properties are
given in this form, we set $\hat{P} = \psi_1$ and $\hat{Q} =\psi_2$,
with the $t$ superscripts omitted for all neurons.  This assumption
can be relaxed significantly; see Sec.~\ref{sec:relaxTimeAgnostic} of
the appendix
%

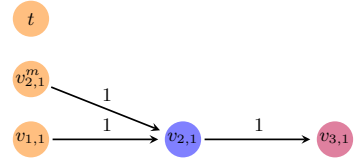
\begin{wrapfigure}[9]{r}{4.5cm} \vspace{-1.2cm}
    \begin{center}
    \scalebox{0.8}{
\begin{tikzpicture}[shorten >=1pt,->,draw=black!50, node distance=\layersep,font=\footnotesize]
  \def\layersep{2.5cm}
    \node[input neuron] (I-1) at (0,-1) {$t$};
    \node[input neuron] (I-2) at (0,-2) {${v}^m_{2,1}$};
    \node[input neuron] (I-3) at (0,-3) {$v_{1,1}$};
    \path[] node[hidden neuron] (H-2)  at (\layersep,-3) {$v_{2,1}$};
    \node[output neuron] at (2*\layersep, -3) (O-2) {$v_{3,1}$};

    \draw[nnedge] (I-2) -- node[above] {$1$} (H-2);
    \draw[nnedge] (I-3) -- node[above] {$1$} (H-2);
    \draw[nnedge] (H-2) -- node[above] {$1$} (O-2);

\end{tikzpicture}

    }
        \caption{
          The feed-forward snapshot network $\hat{N}$ for the RNN from
            Fig.~\ref{fig:rnn_1d}.
        }
    \label{fig:encoding}
    \end{center}
\end{wrapfigure}

\mysubsubsection{Example.}
We demonstrate our approach on the running
example from Fig.~\ref{fig:rnn_1d}. Recall that
$P = \bigwedge_{t=1}^5(-3\leq v^t_{1,1}\leq 3)$,
and
$Q = \bigvee_{t=1}^5(v_{3,1}^t \geq 16)$.
First, we build the snapshot network $\hat{N}$ (Fig.~\ref{fig:encoding})  
by replacing the memory unit $\tilde{v}_{2,1}$ with a regular
neuron, $v^m_{2,1}$, which is connected to node $v_{2,1}$
with weight 1 (the same weight previously found on the edge from
$\tilde{v}_{2,1}$ to $v_{2,1}$), and adding neuron $t$ 
to represent time. Next, we set $\hat{P}$ to be the conjunction of
\begin{inparaenum}[(i)]
\item $P$, with its internal conjunction and $t$ superscripts omitted;
\item the time constraint $1\leq t\leq 5$; and
  \item 
the
invariant that bounds the values of $v^m_{2,1}$ as a function of
time: $v^m_{2,1}\leq 3(t-1)$.
\end{inparaenum}
Our new verification query is thus:
\[
  \langle
  \underbrace{
  v_{1,1} \in \left[-3,3\right]
  \wedge
  t \in \left[1,5\right]
  \wedge
  (v^m_{2,1}\leq 3(t-1))}_{\hat{P}},
  \hat{N},
  \underbrace{v_{3,1}\geq 16}_{\hat{Q}}
  \rangle
\]
This query is, of course, \unsat{}, indicating that the original query is
also \unsat{}. Note that the new node $t$ is added solely for the
purpose of including it in constraints that appear in $\hat{P}$.

The requirement that $I$ be an invariant over the memory units of $N$ 
 ensures that our approach is sound. Specifically, 
it guarantees
that $I$ allows any assignment for $v^m$ that the
original memory unit $\tilde{v}$ might be assigned. 
This is formulated in the following lemma
(whose
proof, by induction, is omitted):
\begin{lemma}
  Let $\varphi = \langle P, N, Q, \tmax\rangle$ be an RNN verification
  query, and let
  $\hat{\varphi} = \langle \hat{P}, \hat{N}, \hat{Q}\rangle$ be the
  snapshot query $\hat{\varphi}=\snapshot{}(\varphi)$.  Specifically,
  let $\hat{P} = P \wedge (1\leq t \leq \tmax)\wedge I$, where $I$ is an
  invariant that bounds the values of each $v^m$.  If $\hat{\varphi}$
  is \unsat{}, then $\varphi$ is also \unsat{}.
\end{lemma}

\subsection{Constructing $\hat{\varphi}:$ Verifying the Invariant}
\label{sec:verifyInvariant}
A key assumption in our reduction from RNN to FFNN verification was
that we were supplied some invariant $I$, which bounds the values of
the $v^m$ neurons as a function of the time $t$. In this section we
make our method more robust, by including a step that verifies that
the supplied formula $I$ is indeed an invariant. This step, too, is
performed by creating an FFNN verification query, which can then be
dispatched using the back-end FFNN verification engine.  (We treat $I$
simultaneously as a formula over the nodes of $\snapshot{}(N)$ and
those of $N$; the translation is performed 
by renaming every occurrence of $v^m$ 
to $\tilde{v}^t$, or vice versa.)

First, we adjust the definitions of an inductive invariant 
(Sec.~\ref{sec:inductive_invariant}) to the RNN setting. 
The state space $Q$ is comprised of states
$ q = \langle \mathcal{A}, t\rangle $, where $\mathcal{A}$ is the
current assignment to the nodes of $N$ (including the assignments of
the memory units), and $t\in\mathbb{N}$ represents time
step. For another state $q'=\langle \mathcal{A}',t'\rangle$, the
transition relation $T(q,q')$ holds if and only if:
\begin{inparaenum}[(i)]
\item $t'=t+1$;  i.e., the time step has advanced by one; 
\item for each neuron $v$ and its memory unit $\tilde{v}$
    it holds that $\mathcal{A}'[\tilde{v}] = \mathcal{A}[v]$; i.e.,
    the vanilla RNN update rule holds; and
\item the assignment $\mathcal{A}'$ of all of the network's neurons
  constitutes a proper evaluation of the RNN according to
  Sec.~\ref{sec:background}; i.e., all weighted sums and activation
  functions are computed properly.
\end{inparaenum}
A state $q_0$ is initial if $t=1$, $\tilde{v} = 0$ for every memory unit,
and the assignment of the network's neurons
constitutes a proper evaluation of the RNN.

Next, let $I$ be a formula over the memory units of $N$, and suppose we
wish to verify that $I$ is an invariant. 
Proving that $I$ is in invariant amounts to proving that 
$I(q_0)$ holds for any initial state $q_0$, and that for every two
states $q,q'\in Q$, $I(q) \wedge T(q,q')
\rightarrow I(q')$. 
Checking whether $I(q_0)$ holds is trivial.
The second check is more tricky; here, the key point is that because
$q$ and $q'$ are consecutive states, the memory units of $q'$ are
simply the neurons of $q$. Thus, we can prove that $I$ holds for $q'$ by looking at
 the snapshot network, assuming that $I$ holds initially, and proving that
$I[v^m\mapsto v, t\mapsto t+1]$, i.e. the invariant with each memory unit
$v^m$ renamed to its corresponding neuron $v$ and the time step
advanced by $1$, also holds. The resulting verification query, which
we term $\varphi_I$, 
 can be verified using the underlying FFNN verification back-end.

We illustrate this process using the running example from Fig.~\ref{fig:rnn_1d}.
Let $I=v^m_{2,1}\leq 3(t-1)$.  $I$ holds at every initial state $q_0$; this is
true because at time $t=1$, $v^m_{2,1} = 0 \leq 3\cdot 0$. Next, we
assume that $I$ holds for state $q=\langle \mathcal{A}, t\rangle$ and
prove that it holds for $q=\langle \mathcal{A}', t + 1\rangle$.
 First, we create the snapshot FFNN $\hat{N}$, shown in
 Fig.~\ref{fig:encoding}. We then extend the original input property
 $P = \bigwedge_{t=1}^5(-3\leq v^t_{1,1}\leq 3)$ into a property
 $P'$ that also captures our assumption that the invariant holds at
 time $t$:
$
   P' = (-3\leq v_{1,1}\leq 3) \wedge (v^m_{2,1}\leq 3(t-1)).
$
 Finally, we prepare an output property $Q'$ that asserts that the
 invariant does not hold for $v_{2,1}$ at time $t+1$, by renaming
 $v_{2,1}^m$ to $v_{2,1}$ and incrementing $t$:
 $
   Q' = \neg( v_{2,1} \leq 3(t+1-1)).
$
When the FFNN verification engine answers
that $\varphi_I=\langle P', \snapshot{}(N), Q'\rangle$ is \unsat{}, we conclude that $I$ is indeed an
invariant. In cases where the query turns out to be \sat{}, $I$ is not
an invariant, and needs to be refined.

Given a formula $I$, the steps described so far allow us to reduce RNN verification to FFNN
verification, in a completely sound and automated way. Next we discuss how to
automate the generation of $I$, as well.
       
\section{Invariant Inference}
\label{sec:invariant_generation}

\subsection{Single Memory Units}
\label{sec:single_dim}

In general, automatic invariant inference 
is undecidable~\cite{PaImShKaSa16}; thus, we employ here a
heuristic approach, that uses \emph{linear templates}.
We first describe the approach on a simple case, in which the network has
a single hidden node $v$ with a memory unit, and then relax this limitation.
Note that the running example depicted in
Fig.~\ref{fig:rnn_1d} fits this case. Here, inferring an invariant according to a
linear template means finding values 
$\alpha_l$ and $\alpha_u$, such that
$
\alpha_l \cdot (t-1) \le \tilde{v}^t \le \alpha_u \cdot (t-1).
$
The value of $\tilde{v}^t$ is thus bounded from below and from above
as a function of time. In our template we use $(t-1)$, and not $t$, in
order to account for the fact that $\tilde{v}^t$ contains the value
that node $v$ was assigned at time $t-1$.  For simplicity, we focus
only on finding the upper bound; the lower bound case is symmetrical.
We have already seen such an upper bound for our running example,
which was sufficiently strong for proving the desired property:
$\tilde{v}^t_{2,1}\leq 3(t-1)$.

Once candidate $\alpha$'s are proposed, verifying that the
invariant holds is performed using the techniques outlined in
Sec.~\ref{sec:method}. There are two places where the process might
fail:
\begin{inparaenum}[(i)]
\item the proposed invariant cannot be proved ($\varphi_I$ is \sat{}), because a
  counter-example exists. This means that our invariant is \emph{too
    strong}, i.e. the bound is too tight. In this case we can weaken
  the invariant by increasing $\alpha_u$; or
\item the proposed invariant holds, but the FFNN verification problem
  that it leads to, $\hat{\varphi}$, is \sat{}. In this case, the invariant is
  \emph{too weak}: it does not
  imply the desired output property. We can
  strengthen the invariant by decreasing $\alpha_u$.
\end{inparaenum}

This search problem leads us to binary search strategy,
described in Alg.~\ref{alg:verification_1d}.  The search stops,
i.e. an optimal invariant is found, when $ub-lb\leq \epsilon$ for a
small constant $\epsilon$. The algorithm
fails if the optimal linear invariant is found, but is insufficient
for proving the property in question; this can happen if $\varphi$ is
indeed \sat{}, or if a more sophisticated invariant is required. 
\begin{algorithm}
\caption{Automatic Single Memory Unit Verification($P,N, Q, \tmax$) }
\begin{algorithmic}[1]
  \State $lb \leftarrow -M$, $ub \leftarrow M$ \Comment{M is a large constant}
  \While {$ub - lb \ge \epsilon$}
      \State $\alpha_{u} \leftarrow \frac{ub+lb}{2},\quad I \leftarrow v_{2,1}^m \le \alpha_u \cdot (t-1)$
      \If {$\varphi_I$ is \unsat{}}
          \State Construct $\hat{\varphi}$ using invariant $I$
          \If {$\hat{\varphi}$ is \unsat{}} 
          \State \Return True
          \EndIf
          \State $ub \leftarrow \alpha_{u}$  \Comment{Invariant too weak}
          \label{step:stronger_invariant}
       \Else
          \State $lb \leftarrow
          \alpha_{u}$ \Comment{Invariant too strong} \label{step:weaker_invariant}
       \EndIf
  \EndWhile
  \State \Return False
\end{algorithmic}
\label{alg:verification_1d}
\end{algorithm}

\mysubsubsection{Discussion: Linear Templates.}
Automated invariant inference has been studied extensively in program
analysis (see Sec.~\ref{sec:relatedWork}). In
particular, elaborate templates have been proposed, which are more
expressive than the linear template that we use. The approach we
presented in Sec.~\ref{sec:method} is general, and is compatible with
many of these templates. Our main motivation for focusing on linear
templates is that most FFNN verification tools readily support linear
constraints, and can thus verify the $\varphi_I$ queries that
originate from linear invariants.  As we demonstrate in
Sec.~\ref{sec:evaluation}, despite their limited expressiveness,
linear invariants are already sufficient for solving many verification
queries.  Extending the technique to work with more expressive
invariants is part of our ongoing work.

\mysubsubsection{Multiple Memory Units in Separate Layers.}
Our approach can be extended to RNNs with multiple memory units,
each in a separate layer, in an iterative
fashion: an invariant is proved separately for each layer, by using
the already-proved invariants of the previous layers. As before, we
begin by constructing the snapshot network in which all memory units
are replaced by regular neurons.  Next, we work layer by layer and
generate invariants that over-approximate each memory unit, by
leveraging the invariants established for memory units in the previous
layers.  Eventually, all memory units are over-approximated using
invariants, and we can attempt to prove the desired property by
solving the snapshot query. An example and the general algorithm for this case
appears in Sec.~\ref{sec:app_multiple_layers} of the appendix
%

\subsection{Layers with Multiple Memory Units}
\label{sec:multidim_verification}

\begin{wrapfigure}[14]{r}{4.5cm}
  \vspace{-1.2cm}
    \begin{center}
    \scalebox{0.8}{ \begin{tabular}{@{}c@{}}

\begin{tikzpicture}[shorten >=1pt,->,draw=black!50, node distance=\layersep,font=\footnotesize]
      \def\layersep{1.5cm}
      \def\vsep{1cm}
      
      \node[input neuron] (I-1) at (0,0) {$v_{1,1}$};
  \path[yshift=0.5] node[shadow neuron] (H-1-s)  at (0.75*\layersep,1.5*\vsep) {$\tilde{v}_{2,1}$};
  \path[yshift=0.5] node[shadow neuron] (H-2-s)  at (0.75*\layersep,-1.5*\vsep) {$\tilde{v}_{2,2}$};
  \path[yshift=0.5] node[hidden neuron] (H-1)  at (2*\layersep,\vsep) {$v_{2,1}$};
  \path[yshift=0.5] node[hidden neuron] (H-2)  at (2*\layersep,-1*\vsep) {$v_{2,2}$};
  \node[output neuron] at (3*\layersep, 0) (O-1) {$v_{3,1}$};

  \draw[nnedge] (I-1) -- node[above, pos=0.3] {-1} (H-1);
  \draw[nnedge] (I-1) -- node[below, pos=0.3] {2} (H-2);
  \draw[nnedge] (H-1-s) -- node[above] {$1$} (H-1);
  \draw[nnedge] (H-1-s) -- node[above, pos=0.2] {\ -1} (H-2);
  \draw[nnedge] (H-1) edge[bend right, dotted, black] node[above] {} (H-1-s);
  \draw[nnedge] (H-2-s) -- node[below] {1} (H-2);
  \draw[nnedge] (H-2-s) -- node[below, pos=0.2] {\ 1} (H-1);
  \draw[nnedge] (H-2) edge[bend left, dotted, black] node[] {} (H-2-s);
  \draw[nnedge] (H-1) -- node[above] {$1$} (O-1);
  \draw[nnedge] (H-2) -- node[above] {$1$} (O-1);

\end{tikzpicture}

\end{tabular}

 }
    \caption{
      An RNN where both memory units
      affect both neurons of the hidden layer: 
        $v^t_{2,1} = \relu{}(\tilde{v}^t_{2,1}  + \tilde{v}^t_{2,2}  
        - v_{1,1}^t)$;
        and
        $v^t_{2,2} = \relu{}(-\tilde{v}^t_{2,1}  + \tilde{v}^t_{2,2}  
        +2  v_{1,1}^t)$.
    }
    \label{fig:rnn_2d}
    \end{center}
\end{wrapfigure}
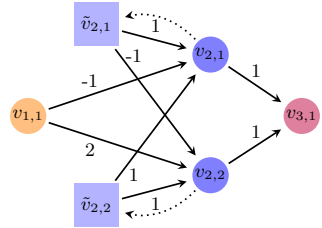

We now extend our approach to support the most general case: an RNN
with layers that contain multiple memory units. We again apply
an iterative, layer-by-layer approach.  The main difficulty is
in inferring invariants for a layer that has multiple memory units, as
in Fig.~\ref{fig:rnn_2d}:  while each memory
unit belongs to a single neuron, it affects the assignments of all
other neurons in that layer.  We propose to handle this case using
separate linear invariants for upper- and lower-bounding each of the
memory units.  However, while the invariants have the same linear form
as in the single memory unit case, \emph{proving} them requires taking
the other invariants of the same layer into account.  Consider the
example in Fig.~\ref{fig:rnn_2d}, and suppose we have
$\alpha_l^1,\alpha_u^1$ and $\alpha_l^2,\alpha_u^2$ for which we wish
to verify that
\begin{equation}
    \alpha_l^1\cdot (t-1) \leq \tilde{v}_{2,1}^t \leq \alpha_u^1\cdot (t-1) \qquad\qquad
    \alpha_l^2\cdot (t-1) \leq  \tilde{v}_{2,2}^t \leq \alpha_u^2\cdot (t-1)
  \label{eq:multidim_bounds}
\end{equation}
In order to prove these bounds we need to dispatch an FFNN verification query
that assumes Eq.~\ref{eq:multidim_bounds} holds and uses it to
prove the inductive step:
\begin{equation}
    \tilde{v}_{2,1}^{t+1} = v_{2,1}^t = \relu{}(- \tilde{v}^t_{1,1} +  \tilde{v}_{2,1}^t + \tilde{v}_{2,2}^t) 
    \leq \alpha_u^1 \cdot (t+1-1)
    \label{eq:multidim_step}
\end{equation}
Similar steps must be performed for $\tilde{v}_{2,1}^{t+1}$'s lower bound, and
also for
$\tilde{v}_{2,2}^{t+1}$'s lower and upper bounds. The key point is that because
Eq.~\ref{eq:multidim_step} involves $\tilde{v}_{2,1}^t$ and $\tilde{v}_{2,2}^t$,
multiple $\alpha$ terms from
Eq.~\ref{eq:multidim_bounds} may need to be used in proving it. This
interdependency means that later changes to some $\alpha$ value
might invalidate previously acceptable assignments for other $\alpha$
values. This adds a layer of complexity that did not exist in the
cases that we had considered previously.

For example, consider the network in Figure~\ref{fig:rnn_2d}, with
$P = \bigwedge_{t=1}^3 -3 \leq v_{1,1}^t \leq 3$,
and
$Q = \bigvee^3_{t=1} v_{3,1}^{t} \ge 100$. 
Our goal is to find values
for  $\alpha_l^1,\alpha_u^1$ and $\alpha_l^2,\alpha_u^2$ that will
satisfy Eq.~\ref{eq:multidim_bounds}. Let us consider 
$\alpha_l^1 =0, \alpha_u^1 = 8, \alpha_l^2 = 0$ and $\alpha_u^2 =
 0$. Using these values, the induction hypothesis
 (Eq.~\ref{eq:multidim_bounds}) and the bounds for $v_{1,1}$,
 we can indeed prove the upper bound for
 $\tilde{v}_{2,1}^{t+1}$:
 \[
     \tilde{v}_{2,1}^{t+1} = v_{2,1}^t = \relu{}(- \tilde{v}^t_{1,1} +  \tilde{v}_{2,1}^t + \tilde{v}_{2,2}^t)
   \leq \relu{}(3 + 8(t-1) +0) \leq 8t
 \]

 Unfortunately, the bounds $0\leq
 \tilde{v}_{2,2}^t\leq 0$ are inadequate, because $\tilde{v}_{2,2}^t$ can
 take on positive values. We are thus required to adjust the $\alpha$
 values, for example by increasing $\alpha_u^2$ to $2$. However, this
 change invalidates the upper bound for $\tilde{v}_{2,1}^{t+1}$,
 i.e. $\tilde{v}_{2,1}^{t+1}\leq 8t$, as that bound relied on the upper bound for
 $\tilde{v}_{2,2}^t$; Specifically, knowing only that $1\leq t \leq 3$, $-3\leq v_{1,1}^t\leq
 3$, $0\leq \tilde{v}_{2,1}^t\leq 8(t-1)$ and $0\leq \tilde{v}_{2,2}^t\leq 2(t-1)$, it is
 impossible to show that $\tilde{v}_{2,1}^{t+1} = v_{2,1}^t  \leq 8t$.


The example above demonstrates the intricate dependencies between the
$\alpha$ values, and the complexity that these dependencies add to the
search process. Unlike in the single memory unit case, it is not immediately clear
how to find an initial invariant that simultaneously holds for all
memory units, or how to strengthen this invariant (e.g., which
$\alpha$ constant to try and improve).

\mysubsubsection{Finding an Initial Invariant.}
We propose to encode the
problem of finding initial $\alpha$ values as a \emph{mixed integer
  linear program} (MILP). The linear and
piecewise-linear constraints that the
$\alpha$ values must satisfy (e.g., Eq.~\ref{eq:multidim_step})
can be precisely encoded in MILP
using standard big-M encoding~\cite{ReluPlex}.
There are two main advantages
to using MILP here:
\begin{inparaenum}[(i)]
\item an MILP solver
  is guaranteed to return a valid invariant, or soundly report that no such
  invariant exists; and
\item MILP instances include a \emph{cost function} to be
  minimized, which can be used to optimize the invariant. For example,
  by setting the cost function to be $ \sum\alpha_u - \sum\alpha_l, $
  the MILP solver will typically generate tight upper and lower bounds.
\end{inparaenum}

The main disadvantage to using MILP is that, in order to ensure that
the invariants hold for all time steps $1\leq t\leq \tmax{}$, we must
encode all of these steps in the MILP query. For example, going back
to Eq.~\ref{eq:multidim_step}, in order to guarantee that 
$v_{2,1}^{t+1} = \relu{}(- v^t_{1,1} +  v_{2,1}^t + v_{2,2}^t) \leq \alpha_u^1 \cdot t$,
we would need to encode within our MILP instance the fact that
$
  \bigwedge_{t=1}^{\tmax}\left( \relu{}(- v^t_{1,1} +  v_{2,1}^t + v_{2,2}^t) \leq \alpha_u^1 \cdot t \right)
$.
This might render the MILP instance difficult to solve for large
values of $\tmax{}$.
However, we stress that this approach is quite
different from, and significantly easier than, unrolling the RNN. The
main reason is that these MILP instances are each generated for a single
layer (as opposed to the entire network in unrolling), which renders
them much simpler. Indeed, in our experiments
(Sec.~\ref{sec:evaluation}), solving these MILP
instances was never the bottleneck.
Still, should this become a problem, we propose to encode
only a subset of the values of $t\in\{1,\ldots,\tmax{}\}$, making the
problem easier to solve; and should the $\alpha$ assignment fail to
produce an invariant (this will be discovered when $\varphi_I$ is
verified), additional constraints could be added to guide the MILP
solver towards a correct solution. We also describe an alternative
approach, which does not require the use of an MILP solver, in
Sec.~\ref{sec:invariantsNoMilp} of the appendix
%

\mysubsubsection{Strengthening the Invariant.}
If we are unable to prove that $\hat{\varphi}$ is
\unsat{} for a given $I$, then the invariant needs to be strengthened. We propose to achieve this by
invoking the MILP solver again, this time adding new linear
constraints for each $\alpha$, that will force the selection of tighter bounds. For example,
if the current invariant is $\alpha_l=3, \alpha_u=7$, we 
add constraints specifying that $\alpha_l\geq 3+\epsilon$ and
$\alpha_u\leq 7-\epsilon$ for some small positive $\epsilon$ ---
leading to stronger invariants. 

\section{Evaluation} 
\label{sec:evaluation}

Our  proof-of-concept implementation of the approach,
called \emph{RnnVerify}, reads an RNN in TensorFlow format. The input
and output properties, $P$ and $Q$, and also $\tmax{}$, are supplied in a simple
proprietary format, and the tool then automatically:
\begin{inparaenum}[(i)]
  \item creates the FFNN snapshot network;
  \item infers a candidate invariant using the MILP heuristics from Sec.~\ref{sec:invariant_generation};
  \item formally verifies that $I$ is an invariant; and
  \item uses $I$ to show that $\hat{\varphi}$, and hence $\varphi$,
    are \unsat{}.
  \end{inparaenum}
If $\hat{\varphi}$ is \sat{}, our module refines $I$ and repeats the
process for a predefined number of steps.

For our evaluation, we focused on neural networks for \emph{speaker
recognition} --- a task for which RNNs are commonly used, because
audio signals tend to have temporal properties and varying lengths. We
applied our verification technique to prove \emph{adversarial
  robustness} properties of these networks, as we describe next.

\mysubsubsection{Adversarial Robustness.}
\label{sec:adversarial_robustness}
It has been shown that alarmingly many neural networks are susceptible to
\emph{adversarial inputs}~\cite{SzZaSuBrErGoFe13}. These inputs are generated by slightly
perturbing correctly-classified inputs, in a way that causes the
misclassification of the perturbed inputs. 
Formally, given a network $N$ that classifies inputs into labels $l_1,\ldots,l_k$, an input $x_0$, and a
target label $l \neq N(x_0)$, an adversarial input is an input
$x$ such that $N(x) = l$ and
$\lVert x-x_0 \rVert\leq\delta$; i.e., input $x$ is very
close to $x_0$, but is misclassified as label $l$.

\emph{Adversarial robustness} is a measure of how difficult it is to
find an adversarial example --- and specifically, what is the smallest
$\delta$ for which such an example 
exists. Verification can be used to find adversarial inputs or rule
out their existence for a given $\delta$, and consequently can find
the smallest $\delta$ for which an adversarial input
exists~\cite{CaKaBaDi17}.  

\mysubsubsection{Speaker Recognition.} A speaker recognition system
receives a voice sample and needs to identify the speaker from a set
of candidates. RNNs are often applied in implementing such
systems~\cite{WaWaPaLo18}, rendering them vulnerable to adversarial
inputs~\cite{KrAdCiKe18}. Because such vulnerabilities in these
systems pose a security concern, it is important to
verify that their underlying RNNs afford high adversarial robustness.

\mysubsubsection{Benchmarks.}
\sloppy
We trained $6$ speaker recognition RNNs, based
on the VCTK dataset~\cite{vctk}. Our networks are of modest, varying
sizes of approximately 220 neurons: they
each contain an input layer of dimension $40$, one or two hidden layers with
$d\in\left\{2,4,8\right\}$ memory units, followed by $5$ fully connected,
memoryless layers with $32$ nodes each, and an output
layer with $20$ nodes. The output nodes represent
the possible speakers between which the RNNs were trained to
distinguish. In addition, in order to enable a comparison to the state
of the art~\cite{AkKeLoPi19}, we trained another, smaller network, which
consists of a single hidden layer. This was
required to accommodate technical constraints in the implementation
of~\cite{AkKeLoPi19}. All networks use ReLUs as their activation functions.

Next, we selected $25$ random, fixed input points
$X = \{x_1,\ldots,x_{25}\}$, that do not change over time; i.e. $x_i\in
\mathbb{R}^{40}$ and $x_i^1=x_i^2=\ldots$ for each $x_i\in X$. Then, 
for each RNN $N$ and input $x_i\in X$, and for each value $2
\leq \tmax \leq 20$, we computed the ground-truth
label $l = N(x_i)$, which is the label that received the highest score
at time step $\tmax{}$. We also computed the label that received the
second-highest score, $l_{sh}$, at time step $\tmax{}$.
Then, for every combination of $N$,
$x_i\in X$, and value of $\tmax{}$, we created the query
$
  \langle
  \bigwedge_{t=1}^{\tmax} (\lVert x'^t - x_i^t \rVert_{L_\infty} \leq 0.01),
  N,
  l_{sh}\geq l
  \rangle
$.
The allowed perturbation, at most $0.01$ in $L_\infty$ norm, was
selected arbitrarily. The query is only \sat{} if there exists an input $x'$ that is at distance at most
$0.01$ from $x$, but for which label $l_{sh}$ is assigned a higher score
than $l$ at time step $\tmax{}$. This formulation resulted in a
total of $2850$ benchmark queries over our 6 networks. 

\begin{wrapfigure}[13]{r}{4.9cm}
   \vspace{-1.2cm}
    \begin{center} 
            \scalebox{.30}{  
                \includegraphics[width=1.3\textwidth]{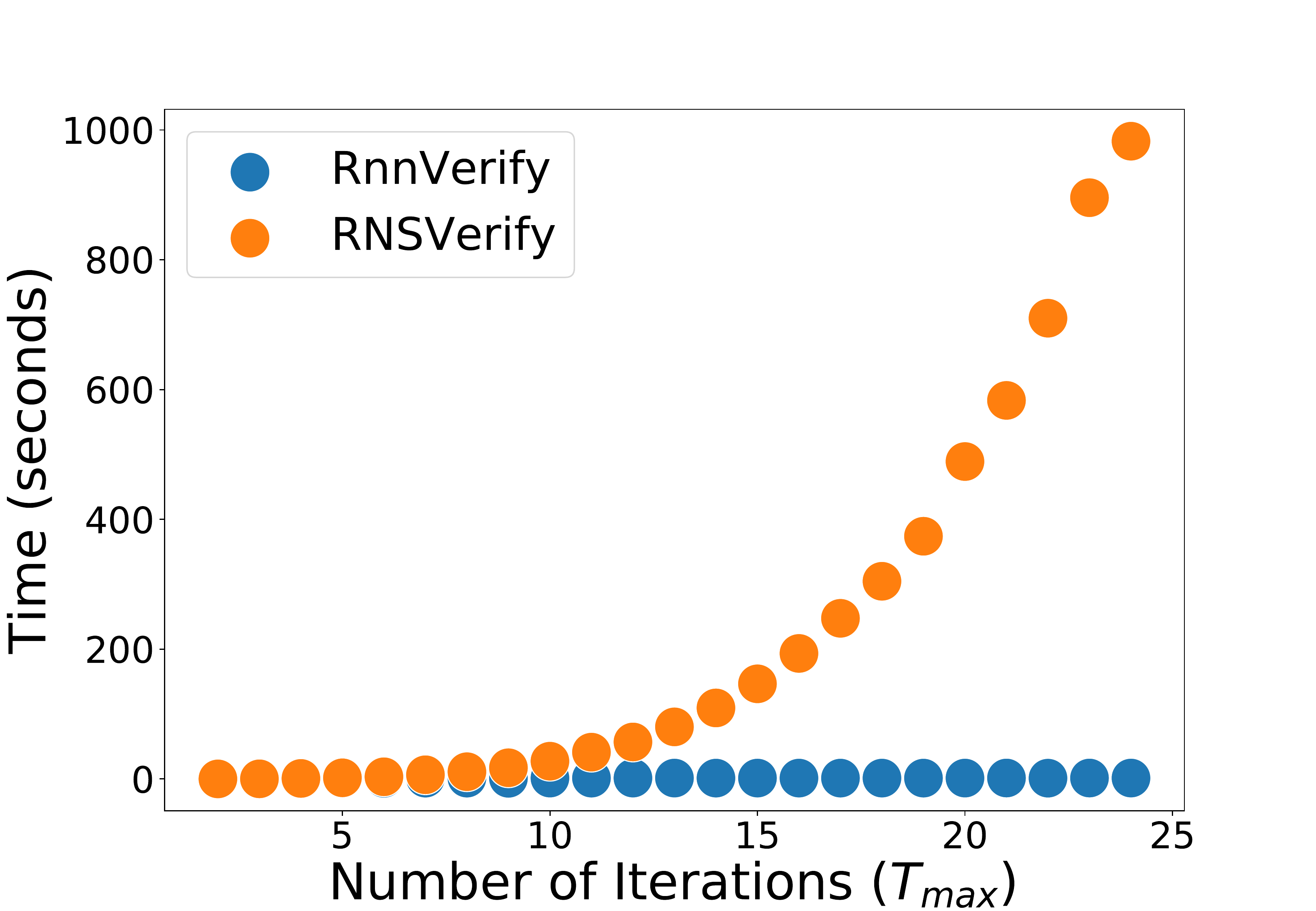}
            }
    \end{center} 
    \caption{ Average running time (in seconds) of RnnVerify and
        RNSVerify, as a function of $\tmax{}$. }
  \label{fig:prev_work_comperssion}
\end{wrapfigure}

\mysubsubsection{Results.}
We began by comparing our technique to the state-of-the-art,
unrolling-based RNSVerify tool~\cite{AkKeLoPi19}, using the small
network we had trained.  Each dot in
Fig.~\ref{fig:prev_work_comperssion} represents a tool's average run
time on the 25 input points, for a specific $\tmax{}$.  Both methods
returned \unsat{} on all queries; however, the runtimes clearly
demonstrate that our approach is far less sensitive to large $\tmax{}$
values.  In a separate experiment, our tool was able to solve a
verification query on the same network with $\tmax{}= 180$ in 2.5
seconds, whereas RNSVerify timed out after 24 hours.

Next, we used RnnVerify on all $2850$ benchmark queries on the 6
larger networks. The results appear in
Sec.~\ref{sec:resultTable} of the appendix
%
and are summarized as follows:
\begin{inparaenum}[(i)]
\item RnnVerify terminated on all benchmarks, with a median runtime
  of 5.39 seconds and an average runtime
  of $48.67$ seconds. The maximal solving time
  was 5701 seconds;
    \item $85\%$ of RnnVerify's runtime was spent within the
      underlying FFNN verification engine, solving $\varphi_I$
      queries. This indicates that as the underlying FFNN verification
      technology improves, our approach will become significantly more scalable;
    \item for $1919$ ($67\%$) of the
        benchmarks, RnnVerify proved that the RNN was robust around the tested point. For
        the remaining $931$ benchmarks, the results are inconclusive: we do not
        know whether the network is vulnerable, or whether more
        sophisticated invariants are needed to prove
        robustness. This demonstrates that for a majority of tested benchmarks, the linear
        template proved useful; and
    \item RnnVerify could generally prove fewer instances with larger values of
      $\tmax$. This is because the linear bounds afforded by our
      invariants become more
      loose as $t$ increases, whereas the neuron's values typically do
      not increase significantly over time. This highlights the need
      for more expressive invariants.
\end{inparaenum}

\section{Related Work} 
\label{sec:relatedWork}

Due to the discovery of undesirable behaviors in many DNNs, multiple
approaches have been proposed for verifying them. These include the
use of SMT solving~\cite{HuKwWaWu17,ReluPlex,Marabou,KuKaGoJuBaKo18},
LP and MILP solving~\cite{Eh17,TjXiTe19}, symbolic interval
propagation~\cite{WaPeWhYaJa18}, abstract
interpretation~\cite{ElGoKa20,GeMiDrTsChVe18}, and many others
(e.g.,~\cite{BuTuToKoMu18,GoFeMaBaKa20,GoKaPaBa18,KaBaKaSc19,LoMa17,NaKaRySaWa17}).
Our technique focuses on RNN verification, but uses an FFNN
verification engine as a back-end. Consequently, it could be
integrated with many of the aforementioned tools, and will benefit
from any improvement in scalability of FFNN verification technology.

Whereas FFNN verification has received a great deal of attention, only
little research has been carried out on RNN verification. Akintunde et
al.~\cite{AkKeLoPi19} were the first to propose such a technique,
based on the notion of unrolling the network into an equivalent FFNN.
Ko et al.~\cite{KoLuWeDaWoLi19} take a different approach, which aims
at quantifying the robustness of an RNN to adversarial inputs ---
which can be regarded as an RNN verification technique tailored for a
particular kind of properties.  The scalability of both approaches is
highly sensitive to the number of time steps, $\tmax{}$, specified by
the property at hand.  In this regard, the main advantage of our
approach is that it is far less sensitive to the number of time steps
being considered.  This affords great
potential for scalability, especially for long sequences of inputs. A
drawback of our approach is that it requires invariant inference,
which is known to be challenging.

In a very recent paper, Zhang et al.~\cite{ZhShGuGuLeNa20} propose a
verification scheme aimed at RNNs that perform cognitive tasks. This
scheme includes computing polytope invariants for the neuron layers of
an RNN, using abstract interpretation and fixed-point analysis. We
consider this as additional evidence of the usefulness of invariant
generation in the context of RNN verification.

Automated invariant inference is a key problem in program
analysis. A few
notable methods for doing so include abstract-interpretation
 (e.g.,~\cite{ShDiDiAi11}); counterexample-guided 
approaches (e.g.,~\cite{NgAnRuHi17}); and learning-based approaches
(e.g.,~\cite{SiDaRaNaSo18}).  It will be interesting to apply these
techniques within the context of our framework, in order to more
quickly and effectively discover useful invariants.

\section{Conclusion} 
\label{sec:conclusion}
Neural network verification is
becoming increasingly important to industry, regulators, and society as
a whole. Research to date has focused primarily on
FFNNs. We propose a novel approach for
the verification of recurrent neural networks --- a kind of
neural networks that is particularly useful for context-dependent
tasks, such as NLP. The cornerstone of our approach is the reduction of
RNN verification to FFNN verification through the use of inductive
invariants. Using a proof-of-concept implementation, we demonstrated
that our approach can tackle many benchmarks orders-of-magnitude
more efficiently than the state of the art. These experiments indicate
the great potential that our approach holds.
In the future, we plan to experiment with more expressive invariants,
and also to apply 
\emph{compositional verification} techniques in order to break the RNN into
multiple, smaller networks, for which invariants can more easily be inferred.

\mysubsubsection{Acknowledgements.}
This work was partially supported by the Semiconductor
Research Corporation, the Binational Science Foundation (2017662), the
Israel Science Foundation (683/18), and the National Science
Foundation (1814369).

\bibliographystyle{abbrv}
\bibliography{rnn}

\begin{thebibliography}{10}

\bibitem{AkKeLoPi19}
M.~Akintunde, A.~Kevorchian, A.~Lomuscio, and E.~Pirovano.
\newblock {Verification of RNN-Based Neural Agent-Environment Systems}.
\newblock In {\em {Proc. 33rd Conf. on Artificial Intelligence (AAAI)}}, pages
  6006--6013, 2019.

\bibitem{BuTuToKoMu18}
R.~Bunel, I.~Turkaslan, P.~Torr, P.~Kohli, and P.~Mudigonda.
\newblock {A Unified View of Piecewise Linear Neural Network Verification}.
\newblock In {\em Proc. 32nd Conf. on Neural Information Processing Systems
  (NeurIPS)}, pages 4795--4804, 2018.

\bibitem{CaKaBaDi17}
N.~Carlini, G.~Katz, C.~Barrett, and D.~Dill.
\newblock {Provably Minimally-Distorted Adversarial Examples}, 2017.
\newblock Technical Report. \url{https://arxiv.org/abs/1709.10207}.

\bibitem{ChNuHuRu18}
C.-H. Cheng, G.~N\"{u}hrenberg, C.-H. Huang, and H.~Ruess.
\newblock {Verification of Binarized Neural Networks via Inter-Neuron
  Factoring}.
\newblock In {\em Proc. 10th Working Conf. on Verified Software: Theories,
  Tools, and Experiments (VSTTE)}, pages 279--290, 2018.

\bibitem{ChNuRu17}
C.-H. Cheng, G.~N\"{u}hrenberg, and H.~Ruess.
\newblock {Maximum Resilience of Artificial Neural Networks}.
\newblock In {\em Proc. 15th Int. Symp. on Automated Technology for
  Verification and Analysis (ATVA)}, pages 251--268, 2017.

\bibitem{CiAdNeKe17}
M.~Cisse, Y.~Adi, N.~Neverova, and J.~Keshet.
\newblock {Houdini: Fooling Deep Structured Visual and Speech Recognition
  Models with Adversarial Examples}.
\newblock In {\em Proc. 30th Advances in Neural Information Processing Systems
  (NIPS)}, pages 6977--6987, 2017.

\bibitem{DeChLeTo18}
J.~Devlin, M.~Chang, K.~Lee, and K.~Toutanova.
\newblock {BERT: Pre-training of Deep Bidirectional Transformers for Language
  Understanding}, 2018.
\newblock Technical Report. \url{http://arxiv.org/abs/1810.04805}.

\bibitem{Eh17}
R.~Ehlers.
\newblock {Formal Verification of Piece-Wise Linear Feed-Forward Neural
  Networks}.
\newblock In {\em Proc. 15th Int. Symp. on Automated Technology for
  Verification and Analysis (ATVA)}, pages 269--286, 2017.

\bibitem{ElGoKa20}
Y.~Elboher, J.~Gottschlich, and G.~Katz.
\newblock {An Abstraction-Based Framework for Neural Network Verification}.
\newblock In {\em Proc. 32nd Int. Conf. on Computer Aided Verification (CAV)},
  2020.

\bibitem{GeMiDrTsChVe18}
T.~Gehr, M.~Mirman, D.~Drachsler-Cohen, E.~Tsankov, S.~Chaudhuri, and
  M.~Vechev.
\newblock {AI2: Safety and Robustness Certification of Neural Networks with
  Abstract Interpretation}.
\newblock In {\em Proc. 39th IEEE Symposium on Security and Privacy (S\&P)},
  2018.

\bibitem{GoFeMaBaKa20}
S.~Gokulanathan, A.~Feldsher, A.~Malca, C.~Barrett, and G.~Katz.
\newblock {Simplifying Neural Networks using Formal Verification}.
\newblock In {\em Proc. 12th NASA Formal Methods Symposium (NFM)}, 2020.

\bibitem{GoAdKeKa20}
B.~Goldberger, Y.~Adi, J.~Keshet, and G.~Katz.
\newblock {Minimal Modifications of Deep Neural Networks using Verification}.
\newblock In {\em Proc. 23rd Int. Conf. on Logic for Programming, Artificial
  Intelligence and Reasoning (LPAR)}, pages 260--278, 2020.

\bibitem{dnn}
I.~Goodfellow, Y.~Bengio, and A.~Courville.
\newblock {\em {Deep Learning}}.
\newblock MIT Press, 2016.

\bibitem{GoKaPaBa18}
D.~Gopinath, G.~Katz, C.~P\v{a}s\v{a}reanu, and C.~Barrett.
\newblock {DeepSafe: A Data-driven Approach for Assessing Robustness of Neural
  Networks}.
\newblock In {\em Proc. 16th. Int. Symposium on on Automated Technology for
  Verification and Analysis (ATVA)}, pages 3--19, 2018.

\bibitem{HuKwWaWu17}
X.~Huang, M.~Kwiatkowska, S.~Wang, and M.~Wu.
\newblock {Safety Verification of Deep Neural Networks}.
\newblock In {\em Proc. 29th Int. Conf. on Computer Aided Verification (CAV)},
  pages 3--29, 2017.

\bibitem{RNNCode}
Y.~Jacoby, C.~Barrett, and G.~Katz.
\newblock {RnnVerify}, 2020.
\newblock \url{https://github.com/yuvaljacoby/RnnVerify}.

\bibitem{ReluPlex}
G.~Katz, C.~Barrett, D.~Dill, K.~Julian, and M.~Kochenderfer.
\newblock {Reluplex: An Efficient {SMT} Solver for Verifying Deep Neural
  Networks}.
\newblock In {\em Proc. 29th Int. Conf. on Computer Aided Verification (CAV)},
  pages 97--117, 2017.

\bibitem{KaBaDiJuKo17}
G.~Katz, C.~Barrett, D.~Dill, K.~Julian, and M.~Kochenderfer.
\newblock {Towards Proving the Adversarial Robustness of Deep Neural Networks}.
\newblock In {\em Proc. 1st Workshop on Formal Verification of Autonomous
  Vehicles, (FVAV)}, pages 19--26, 2017.

\bibitem{Marabou}
G.~Katz, D.~Huang, D.~Ibeling, K.~Julian, C.~Lazarus, R.~Lim, P.~Shah,
  S.~Thakoor, H.~Wu, A.~Zelji{\'{c}}, D.~Dill, M.~Kochenderfer, and C.~Barrett.
\newblock {The Marabou Framework for Verification and Analysis of Deep Neural
  Networks}.
\newblock In {\em Proc. 31st Int. Conf. on Computer Aided Verification (CAV)},
  pages 443--452, 2019.

\bibitem{KaBaKaSc19}
Y.~Kazak, C.~Barrett, G.~Katz, and M.~Schapira.
\newblock {Verifying Deep-RL-Driven Systems}.
\newblock In {\em Proc. 1st ACM SIGCOMM Workshop on Network Meets AI \& ML
  (NetAI)}, pages 83--89, 2019.

\bibitem{KoLuWeDaWoLi19}
C.~Ko, Z.~Lyu, T.~Weng, L.~Daniel, N.~Wong, and D.~Lin.
\newblock {POPQORN: Quantifying Robustness of Recurrent Neural Networks}.
\newblock In {\em Proc. 36th IEEE Int. Conf. on Machine Learning and
  Applications (ICML)}, 2019.

\bibitem{KrAdCiKe18}
F.~Kreuk, Y.~Adi, M.~Cisse, and J.~Keshet.
\newblock {Fooling End-to-End Speaker Verification with Adversarial Examples}.
\newblock In {\em Proc. IEEE Int. Conf. on Acoustics, Speech and Signal
  Processing (ICASSP)}, pages 1962--1966, 2018.

\bibitem{KuKaGoJuBaKo18}
L.~Kuper, G.~Katz, J.~Gottschlich, K.~Julian, C.~Barrett, and M.~Kochenderfer.
\newblock {Toward Scalable Verification for Safety-Critical Deep Networks},
  2018.
\newblock Technical Report. \url{https://arxiv.org/abs/1801.05950}.

\bibitem{LiKaElWe16}
Z.~Lipton, D.~Kale, C.~Elkan, and R.~Wetzel.
\newblock {Learning to Diagnose with LSTM Recurrent Neural Networks}.
\newblock In {\em Proc. 4th Int. Conf. on Learning Representations (ICLR)},
  2016.

\bibitem{LoMa17}
A.~Lomuscio and L.~Maganti.
\newblock {An Approach to Reachability Analysis for Feed-Forward ReLU Neural
  Networks}, 2017.
\newblock Technical Report. \url{http://arxiv.org/abs/1706.07351}.

\bibitem{NaKaRySaWa17}
N.~Narodytska, S.~Kasiviswanathan, L.~Ryzhyk, M.~Sagiv, and T.~Walsh.
\newblock {Verifying Properties of Binarized Deep Neural Networks}, 2017.
\newblock Technical Report. \url{http://arxiv.org/abs/1709.06662}.

\bibitem{NgAnRuHi17}
T.~Nguyen, T.~Antonopoulos, A.~Ruef, and M.~Hicks.
\newblock {Counterexample-Guided Approach to Finding Numerical Invariants}.
\newblock In {\em Proc. 11th Joint Meeting on Foundations of Software
  Engineering (FSE)}, pages 605--615, 2017.

\bibitem{PaImShKaSa16}
O.~Padon, N.~Immerman, S.~Shoham, A.~Karbyshev, and M.~Sagiv.
\newblock {Decidability of Inferring Inductive Invariants}.
\newblock In {\em Proc. 43th Symposium on Principles of Programming Languages
  (POPL)}, pages 217--231, 2016.

\bibitem{ShDiDiAi11}
R.~Sharma, I.~Dillig, T.~Dillig, and A.~Aiken.
\newblock {Simplifying Loop Invariant Generation Using Splitter Predicates}.
\newblock In {\em Proc. 23rd Int. Conf. on Computer Aided Verification (CAV)},
  pages 703--719, 2011.

\bibitem{SiDaRaNaSo18}
X.~Si, H.~Dai, M.~Raghothaman, M.~Naik, and L.~Song.
\newblock {Learning Loop Invariants for Program Verification}.
\newblock In {\em Proc. 32nd Conf. on Neural Information Processing Systems
  (NeurIPS)}, pages 7762--7773, 2018.

\bibitem{SzZaSuBrErGoFe13}
C.~Szegedy, W.~Zaremba, I.~Sutskever, J.~Bruna, D.~Erhan, I.~Goodfellow, and
  R.~Fergus.
\newblock {Intriguing Properties of Neural Networks}, 2013.
\newblock Technical Report. \url{http://arxiv.org/abs/1312.6199}.

\bibitem{TjXiTe19}
V.~Tjeng, K.~Xiao, and R.~Tedrake.
\newblock {Evaluating Robustness of Neural Networks with Mixed Integer
  Programming}.
\newblock In {\em Proc. 7th Int. Conf. on Learning Representations (ICLR)},
  2019.

\bibitem{WaWaPaLo18}
L.~Wan, Q.~Wang, A.~Papir, and I.~Lopez{-}Moreno.
\newblock {Generalized End-to-End Loss for Speaker Verification}, 2017.
\newblock Technical Report. \url{http://arxiv.org/abs/1710.10467}.

\bibitem{WaPeWhYaJa18}
S.~Wang, K.~Pei, J.~Whitehouse, J.~Yang, and S.~Jana.
\newblock {Formal Security Analysis of Neural Networks using Symbolic
  Intervals}.
\newblock In {\em Proc. 27th {USENIX} Security Symposium}, pages 1599--1614,
  2018.

\bibitem{WuOzZeIrJuGoFoKaPaBa20}
H.~Wu, A.~Ozdemir, A.~Zelji\'c, A.~Irfan, K.~Julian, D.~Gopinath, S.~Fouladi,
  G.~Katz, C.~P\u{a}s\u{a}reanu, and C.~Barrett.
\newblock {Parallelization Techniques for Verifying Neural Networks}, 2020.
\newblock Technical Report. \url{https://arxiv.org/abs/2004.08440}.

\bibitem{vctk}
J.~Yamagishi, C.~Veaux, and K.~MacDonald.
\newblock {CSTR VCTK Corpus: English Multi-speaker Corpus for CSTR Voice
  Cloning Toolkit}, 2019.
\newblock University of Edinburgh. \url{https://doi.org/10.7488/ds/2645}.

\bibitem{ZhShGuGuLeNa20}
H.~Zhang, M.~Shinn, A.~Gupta, A.~Gurfinkel, N.~Le, and N.~Narodytska.
\newblock {Verification of Recurrent Neural Networks for Cognitive Tasks via
  Reachability Analysis}.
\newblock In {\em Proc. 24th Conf. of European Conference on Artificial
  Intelligence (ECAI)}, 2020.

\end{thebibliography}

\newpage
\noindent
\textbf{\Huge Appendix}

\section{Time-Dependent Properties}
\label{sec:relaxTimeAgnostic}

Our technique for verifying inferred invariants and for using them to
solve the snapshot query (and hence to prove the property in question)
hinges on our ability to reduce each step into an FFNN verification
query. In order to facilitate this, we made the simplifying assumption
that properties $P$ and
  $Q$ of the RNN verification query  are time-agnostic; i.e. they are of the form
  $P = \bigwedge_{t=1}^{\tmax}\psi_1$ and
  $Q=\bigvee_{t=1}^{\tmax}\psi_2$ for $\psi_1$ and $\psi_2$ that are
  conjunctions of linear constraints. However, this limitation 
can be relaxed significantly.

Currently, input property $P$ specifies a constant range for the
inputs, e.g. $-3 \leq v_{1,1}^t \leq 3$ for all $1\leq
t\leq\tmax{}$. However, we observe that our technique can be applied
also for input properties that encode linear time constraints;
e.g. $4t\leq v_{1,1}^t\leq 5t$. These properties can be transferred,
as-is, to the FFNN snapshot network, and are compatible with our
proposed technique. Likewise, the output property $Q$ can also include
constraints that are linear in $t$; and can also restrict the query
to a particular time step $t=t_0$.  In fact, even more complex,
piecewise-linear constraints can be encoded and are compatible with
our technique. However, encoding these constraints might entail adding
additional neurons to the RNN. For example, the constraint
$(\max(v_{1,1}^t,v_{1,2}^t)\geq 5t)$ is piecewise-linear and can be
encoded~\cite{CaKaBaDi17}; the encoding itself is technical, and is
omitted. As for constraints that are not piecewise-linear, if these
can be soundly approximated using piecewise-linear constraints, then
they can be soundly handled using our technique.

\section{Verification Algorithm for RNNs with Multiple Memory Units in Separate Layers}
\label{sec:app_multiple_layers}

As described in Sec.~\ref{sec:single_dim}, we take an iterative
approach to verify networks with multiple memory units, each in a
separate layer. We start from the input layer and prove an invariant
for each layer that has memory units; and when proving an invariant
for layer $i+1$, we use the already-proven invariants of layers $1$
through $i$.

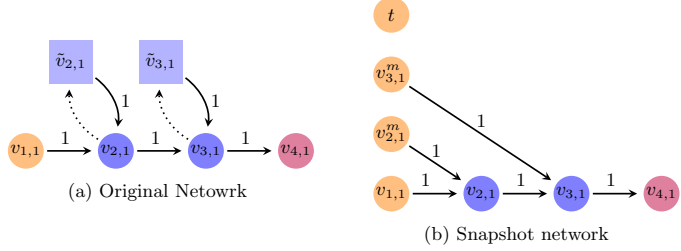
\begin{figure}[htp]
\begin{center}
\scalebox{0.79}{
    \begin{tabular}{@{}c@{}}

\begin{subfigure}[h]{.42\textwidth}
    \centering\begin{tikzpicture}[shorten >=1pt,->,draw=black!50, node distance=\layersep,font=\footnotesize]
          \def\layersep{1.5cm}
      \node[input neuron] (I-1) at (0,0) {$v_{1,1}$};
      \path[] node[hidden neuron] (H-1)  at (\layersep,0) {$v_{2,1}$};
      \path[] node[shadow neuron] (H-1-s)  at (0.5*\layersep,1.5) {$\tilde{v}_{2,1}$};

      \path[] node[hidden neuron] (H-2)  at (2*\layersep,0) {$v_{3,1}$};
      \path[] node[shadow neuron] (H-2-s)  at (1.5*\layersep,1.5) {$\tilde{v}_{3,1}$};
      \node[output neuron] at (3*\layersep, 0) (O-1) {$v_{4,1}$};

        \draw[nnedge] (H-1-s) edge[bend left, black] node[right] {$1$} (H-1);
        \draw[nnedge] (H-1) edge[bend left, dotted,black] node[] {} (H-1-s);

        \draw[nnedge] (H-2-s) edge[bend left, black] node[right] {$1$} (H-2);
        \draw[nnedge] (H-2) edge[bend left, dotted,black] node[] {} (H-2-s);
        \draw[nnedge] (I-1) -- node[above, pos=0.4] {$1$} (H-1);
        \draw[nnedge] (H-1) -- node[above, pos=0.4] {$1$} (H-2);
        \draw[nnedge] (H-2) -- node[above, pos=0.4] {$1$} (O-1);

    \end{tikzpicture}
    \caption{Original Netowrk
        \label{fig:2l_rnn_a}
    }
\end{subfigure}
~~~~~~~




\begin{subfigure}[h]{.4\textwidth}
    \begin{tikzpicture}[shorten >=1pt,->,draw=black!50, node distance=\layersep,font=\footnotesize]
          \def\layersep{1.5cm}
          \def\vsep{1cm}
      \node[input neuron] (I-1) at (0,0) {$v_{1,1}$};
      \path[] node[input neuron] (T)  at (0,3*\vsep) {$t$};
      \path[] node[input neuron] (H-1_new)  at (0,\vsep) {$v^m_{2,1}$};
      \path[] node[hidden neuron] (H-1)  at (\layersep,0) {$v_{2,1}$};
      \path[] node[hidden neuron] (H-2)  at (2*\layersep,0) {$v_{3,1}$};
      \path[] node[input neuron] (H-2_new)  at (0,2*\vsep) {$v^m_{3,1}$};
      \node[output neuron] at (3*\layersep, 0) (O-1) {$v_{4,1}$};

      \draw[nnedge] (I-1) -- node[above, pos=0.3] {$1$} (H-1);
        \draw[nnedge] (H-1_new) -- node[above] {$1$} (H-1);
        \draw[nnedge] (H-2_new) -- node[above] {$1$} (H-2);
        \draw[nnedge] (H-1) -- node[above, pos=0.4] {$1$} (H-2);
        \draw[nnedge] (H-2) -- node[above, pos=0.4] {$1$} (O-1);

    \end{tikzpicture}
    \caption{Snapshot network
        \label{fig:2l_rnn_c}
    }
\end{subfigure}

\end{tabular}

}
    \caption{
      An RNN with multiple memory units, in
      separate layers (on the left), and its snapshot network (on the right). 
    }

\label{fig:2l_rnn}
\end{center}
\end{figure}

An example appears in Fig.~\ref{fig:2l_rnn}. Let
$P = \bigwedge_{t=1}^5(-3\leq v^t_{1,1}\leq 3)$ and
$Q = \bigvee_{t=1}^5(v_{4,1}^t \geq 60)$.  First we construct the
snapshot network shown in the figure.  Next, we prove the invariant
$v_{2,1}^m\leq 3(t-1)$, same as we did before. This invariant bounds
the values of $v_{2,1}^m$. Next, we use this information in proving an
invariant also for $v_{3,1}$; e.g., $v_{3,1}^m \le 9\cdot (t-1)$. To
see why this is an invariant, observe that
\[
  v_{3,1}^t \leq \sum_{i=1}^tv_{2,1}^i \stackrel{*}{\leq} \sum_{i=1}^t 3i =
(3+3t)\frac{t}{2} \leq (3+3\tmax)\frac{t}{2} = 9t
\]

Note that the invariant for $v_{2,1}$ was used in the $*$ transition.  Once this
second invariant is proved, we can show that the original property holds, by
using FFNN verification to show that 
$
  \langle
  P,
  \snapshot(N),
  v_{4,1}\geq 60
  \rangle
$
is \unsat{}; where $\snapshot(N)$ is the FFNN from Fig.~\ref{fig:2l_rnn}, and
\[
  P = (-3\leq v_{1,1}\leq 3) \wedge (1\leq t \leq 5) \wedge (v_{2,1}^m\leq
     3(t-1))  \wedge (v_{3,1}^m\leq 9(t-1))
   \]

Algorithm~\ref{alg:multi_layer_verification} describe the complete process.
We assume for simplicity that \emph{every} hidden layer in the RNN has a
(single) memory unit. Initially
(Lines~\ref{line:initialInvariantsStart}-\ref{line:initialInvariantsEnd}),
the algorithm guesses and verifies a very coarse upper bound on each
of the memory units. Next, it repeatedly attempts to solve the
snapshot query $\hat{\varphi}$ using the current invariants. If
successful, we are done (Line~\ref{line:success}); and otherwise, we
start another pass over the network's layers, attempting to strengthen
each invariant in turn. We know we have reached an optimal invariant
for a layer when the search range for that layer's $\alpha$ becomes
smaller than some small $\epsilon$ constant
(Line~\ref{line:rangeTooSmall}). The algorithm fails
(Line~\ref{line:failure}) when (i) optimal invariants for all layers
have been discovered; and (ii) these optimal invariants are
insufficient for solving the snapshot query.  For simplicity, the algorithm
only deals with upper bounds, but it can be extended to incorporate lower
bound invariants as well in a straightforward manner.
 
\begin{algorithm}[h]
  \caption{Automatic Multiple Memory Units Verification($P,N,Q,\tmax$)}
  \begin{algorithmic}[1]
    \State Construct the snapshot network $\hat{N}$
    \For {$i=2$ to $n-1$} \label{line:initialInvariantsStart}
    \State $lb_i\leftarrow -M\cdot i, ub_i\leftarrow M\cdot i, \alpha_i\leftarrow M\cdot i$ \Comment{M is a large constant}

    \State $I_i\leftarrow v_i^m\leq \alpha_i \cdot t$ \Comment{Loose invariant, should hold}
    \If {$\varphi_{I_i}$ is \sat{}}
    \Return False \Comment{Loose invariant fails, give up}
    \EndIf
    \EndFor \label{line:initialInvariantsEnd}
    \While {True}
    \State Construct $\hat{\varphi}$ using the invariant $ \bigwedge_{i=2}^{n-1}I_i$
    \If {$\hat{\varphi}$ is \unsat{}}
    \State \Return True \Comment{Invariants sufficiently strong} \label{line:success}
    \Else
      \State progressMade $\leftarrow$ False
      \For {$i=2$ to $n-1$}
      \If {$ub_i - lb_i \leq \epsilon$} \label{line:rangeTooSmall}
      \State Continue \Comment{Already have optimal invariant}
      \EndIf
      \State progressMade $\leftarrow$ True \Comment{Still searching for
      optimal invariant}
      \State $\alpha_i'\leftarrow (ub_i + lb_i) / 2$
      \State $I_i'\leftarrow v_i^m\leq \alpha_i' \cdot t$
      \If {$\varphi_{I_i}'$ is \unsat{}}
      \State $I_i\leftarrow I_i',\alpha_i\leftarrow \alpha_i',ub_i\leftarrow \alpha_i$ \Comment{Better invariant found}
      \Else
      \State $lb_i\leftarrow \alpha_i'$ \Comment{Invariant unchanged,
      range shrinks}
      \EndIf
      \EndFor
      \If {!progressMade}
       \Return False\ \Comment{Best invariants too weak} \label{line:failure}
      \EndIf
      \EndIf
      \EndWhile
     \end{algorithmic}
  \label{alg:multi_layer_verification}
\end{algorithm}

A key property of our layer-by-layer approach for invariant inference is that
accurate invariants for layer $i$ are crucial for proving invariants for layer
$i+1$. When each layer contains a single neuron, finding the optimal invariant
for each layer is straightforward; but, as we discuss in
Sec.~\ref{sec:multidim_verification}, things are not as simple when multiple
memory units reside in the same layer.

\FloatBarrier

\section{Invariant Inference without MILP}
\label{sec:invariantsNoMilp}

In cases where using an MILP solver is undesirable (for example, if
$\tmax$ is very large and the MILP instances become a bottleneck), we
propose an \emph{incremental approach}. Here, we start
with an arbitrary assignment of $\alpha$ values, which may or may not
constitute an invariant, and iteratively change one $\alpha$ at a time.
This change needs to be ``in the right direction'' ---
i.e., if our $\alpha$'s do not currently constitute an invariant, we
need to weaken the bounds; and if they do constitute an invariant but
that invariant is too weak to solve $\hat{\varphi}$, we need to
tighten the bounds. The selection of which $\alpha$ to change and by
how much to change it can be random, arbitrary, or according to some
heuristic.  In our experiments we observed that while this approach is
computationally cheaper than solving MILP instances, it tends to lead
to longer sequences of refining the $\alpha$'s before an appropriate
invariant is found. Devising heuristics that will improve the
efficiency of this approach remains a topic for future work.

\section{Experimental Results}
\label{sec:resultTable}

Table~\ref{table:results} fully describes the result of running
RnnVerify on all $2850$ benchmark queries that we used.  Network
$N_{x,y}$ has a hidden layer with $x$ memory units; a second hidden
layer with $y$ memory units, if $y>0$; and $5$ fully connected layers,
each with $32$ nodes.  All experiments were run using using an Intel
Xeon E5 machine with $6$ cores and $3$GB of memory.  Each entry
depicts the average runtime, in seconds, over the 25 input points; and
also the number $x/25$ of queries where RnnVerify successfully proved
adversarial robustness. In the remaining $25-x$ queries, linear
invariants were insufficient for proving that the snapshot query is
\unsat{}. The results are generally monotonic: as $\tmax{}$ increases,
fewer instances can be verified. We used $\dagger$ to mark the few
cases where monotonicity was broken. Our inspection revealed that
these cases were due to \emph{numerical instability} that occurred in
either the underlying MILP solver or the FFNN verification tools.
  
\begin{table}[htp]
\centering
\caption{
  Running RnnVerify on our $2850$ benchmarks. 
}
\begin{tabular}[htp]{rc|clc|clc|clc|clc|clc|clc}
  \toprule
  $\tmax{}$ &&&
  \multicolumn{1}{c}{$N_{2,0}$} &&&
  \multicolumn{1}{c}{$N_{2,2}$} &&&
  \multicolumn{1}{c}{$N_{4,0}$} &&&
  \multicolumn{1}{c}{$N_{4,2}$} &&&
  \multicolumn{1}{c}{$N_{4,4}$} &&&
  \multicolumn{1}{c}{$N_{8,0}$} &
    \\
  \midrule
	2 &&&
0.21(25/25) &&& 0.69(25/25) &&& 0.56(25/25) &&& 0.60(25/25) &&& 1.13(25/25) &&& 1.80(25/25) &
	\\
	3 &&&
4.00(25/25) &&& 21.72(25/25) &&& 17.06(25/25) &&& 39.05(25/25) &&& 21.42(25/25) &&& 28.82(5/25) &
	\\
	4 &&&
4.94(25/25) &&& 28.54(25/25) &&& 27.52(25/25) &&& 80.18(23/25) &&& 42.04(25/25) &&& 9.08(2/25) &
	\\
	5 &&&
5.57(25/25) &&& 30.01(25/25) &&& 24.22(25/25) &&& 48.97(19/25) &&& 11.61(25/25) &&& 12.61(2/25) &
	\\
	6 &&&
6.01(25/25) &&& 22.92(25/25) &&& 57.14(24/25)\textdagger &&& 78.09(17/25) &&& 4.80(25/25) &&& 18.01(2/25) &
	\\
	7 &&&
6.39(25/25) &&& 31.43(25/25) &&& 61.79(25/25) &&& 68.70(17/25) &&& 3.03(9/25) &&& 19.43(2/25) &
	\\
	8 &&&
6.64(25/25) &&& 7.80(25/25) &&& 47.60(25/25) &&& 98.05(15/25)\textdagger &&& 3.07(8/25)\textdagger &&& 27.65(2/25) &
	\\
	9 &&&
7.05(25/25) &&& 8.66(25/25) &&& 85.26(23/25) &&& 147.65(16/25) &&& 3.73(9/25) &&& 26.49(2/25) &
	\\
	10 &&&
7.06(25/25) &&& 10.72(25/25) &&& 82.29(14/25) &&& 239.96(16/25) &&& 3.99(8/25)\textdagger &&& 33.12(2/25) &
	\\
	11 &&&
7.38(25/25) &&& 13.38(25/25) &&& 122.94(14/25) &&& 150.03(16/25) &&& 4.71(8/25)\textdagger &&& 29.59(2/25) &
	\\
	12 &&&
7.46(25/25) &&& 10.28(25/25) &&& 107.15(14/25) &&& 15.75(16/25) &&& 4.33(8/25)\textdagger &&& 47.05(2/25) &
	\\
	13 &&&
14.61(25/25) &&& 10.61(25/25) &&& 218.79(13/25) &&& 16.93(16/25) &&& 4.87(9/25) &&& 43.38(2/25) &
	\\
	14 &&&
7.95(25/25) &&& 11.41(25/25) &&& 149.37(12/25) &&& 18.08(16/25) &&& 5.41(8/25) &&& 44.16(2/25) &
	\\
	15 &&&
14.50(25/25) &&& 11.99(25/25) &&& 365.37(12/25) &&& 19.56(16/25) &&& 6.12(8/25) &&& 52.22(2/25) &
	\\
	16 &&&
8.24(25/25) &&& 12.56(25/25) &&& 294.83(12/25) &&& 20.71(16/25) &&& 6.67(8/25) &&& 55.62(2/25) &
	\\
	17 &&&
8.30(25/25) &&& 13.30(25/25) &&& 323.97(12/25) &&& 20.90(15/25) &&& 6.65(8/25) &&& 66.14(2/25) &
	\\
	18 &&&
8.35(25/25) &&& 13.93(25/25) &&& 481.18(12/25) &&& 21.68(15/25) &&& 7.71(8/25) &&& 82.93(2/25) &
	\\
	19 &&&
8.56(25/25) &&& 14.46(25/25) &&& 350.61(12/25) &&& 23.36(15/25) &&& 7.85(8/25) &&& 75.68(2/25) &
	\\
	20 &&&
8.61(25/25) &&& 15.17(25/25) &&& 264.39(12/25) &&& 24.28(15/25) &&& 8.24(8/25) &&& 71.26(2/25) &
	\\
  \midrule
  \bottomrule
\end{tabular}%
\label{table:results}
\end{table}%

\end{document}